\documentclass[conference]{IEEEtran}
\IEEEoverridecommandlockouts
\usepackage{cite}
\usepackage[numbers,sort&compress]{natbib}
\usepackage{hyperref}
\usepackage{amsmath,amssymb,amsfonts}
\usepackage{algorithmic}
\usepackage{graphicx}
\usepackage{textcomp}
\usepackage{xcolor}
\usepackage{multirow}
\usepackage{array}
\usepackage{makecell}
\usepackage{subcaption}
\usepackage{cleveref}
\crefname{figure}{fig.}{figures}
\Crefname{figure}{Fig.}{Figures}
\usepackage{svg}
\usepackage{caption}
\usepackage{booktabs} 
\usepackage{threeparttable}

\def\BibTeX{{\rm B\kern-.05em{\sc i\kern-.025em b}\kern-.08em
    T\kern-.1667em\lower.7ex\hbox{E}\kern-.125emX}}
\begin{document}

\title{Multivariate Wireless Link Quality Prediction Based on Pre-trained Large Language Models\\
}

\author{\IEEEauthorblockN{Zhuangzhuang Yan$^{1}$, Xinyu Gu$^{1,2}$, Shilong Fan$^{1}$, Zhenyu Liu$^{3}$}
\IEEEauthorblockA{\textit{$^{1}$ Beijing University of Posts and Telecommunications, }Beijing , China \\
\textit{$^{2}$Purple Mountain Laboratories,} Nanjing , China \\
\textit{$^{3}$Institute for Communication Systems, University of Surrey,} United Kingdom, Guildford \\
Email: yanzz,  guxinyu, fansl@bupt.edu.cn, zhenyu.liu@surrey.ac.uk
}
}

\sloppy
\maketitle

\begin{abstract}
Accurate and reliable link quality prediction (LQP) is crucial for optimizing network performance, ensuring communication stability, and enhancing user experience in wireless communications. However, LQP faces significant challenges due to the dynamic and lossy nature of wireless links, which are influenced by interference, multipath effects, fading, and blockage. In this paper, we propose GAT-LLM, a novel multivariate wireless link quality prediction model that combines Large Language Models (LLMs) with Graph Attention Networks (GAT) to enable accurate and reliable multivariate LQP of wireless communications. By framing LQP as a time series prediction task and appropriately preprocessing the input data, we leverage LLMs to improve the accuracy of link quality prediction. To address the limitations of LLMs in multivariate prediction due to typically handling one-dimensional data, we integrate GAT to model interdependencies among multiple variables across different protocol layers, enhancing the model's ability to handle complex dependencies. Experimental results demonstrate that GAT-LLM significantly improves the accuracy and robustness of link quality prediction, particularly in multi-step prediction scenarios.
\end{abstract}


\section{Introduction}
\IEEEPARstart{A}{s} the application scenarios of wireless networks become increasingly diverse and complex, the quality of wireless links significantly impacts network performance, communication reliability, and user experience. Therefore, efficiently and accurately predicting link quality to maintain network stability and enhance user experience has become a pressing challenge in wireless communications. However, in wireless communication networks, factors such as interference, multipath effects, fading, and blockage can cause fluctuations in wireless signals, resulting in unstable and time-varying wireless links. Additionally, radio propagation conditions vary significantly over time and space \cite{ref2}, further complicating link quality prediction (LQP).


Exising LQP traditional methods primarily relied on classical statistical models. For example, an autoregressive integrated moving average (ARIMA) model was applied in \cite{ref-arima} to reduce channel congestion through link quality prediction. With the advent of deep learning, more advanced methods have emerged. In \cite{ref3-1}, the authors utilized a convolutional long short-term memory (Conv-LSTM) model to predict the link quality in point cloud-based millimeter-wave communication systems. \cite{ref4} proposed a temporal convolutional network based on an improved self-attention mechanism (TCNS), where a self-attention mechanism (SAM) is employed to capture short-term correlations in link quality, thereby enhancing prediction accuracy. Although deep learning-based models have demonstrated strong performance in predicting time series for wireless link quality, they also encounter limitations and challenges. For instance, these models typically require a mass of training data to perform effectively. In certain application scenarios, acquiring sufficient data can be problematic. Additionally, deep learning models may exhibit instability in their predictions when exposed to noise or outliers, compromising the model’s robustness and overall reliability.

To address these challenges, large language models (LLMs) offer a promising alternative. Unlike traditional deep learning models, LLMs possess strong contextual understanding and generalization capabilities, enabling them to perform effectively with relatively limited training data \cite{ref5}. Moreover, LLMs excel at processing complex sequential data and capturing long-term dependencies, leveraging their advanced pattern recognition and reasoning capabilities to facilitate efficient time series prediction \cite{ref6, ref7}. In \cite{ref6}, the authors developed a foundational model for time series analysis, demonstrating strong performance in time series prediction tasks. \cite{ref7} introduced a novel framework called Time-LLM, which repurposes LLMs for general time series prediction while maintaining the underlying integrity of the original language models. Despite growing evidence that large models hold the potential to revolutionize wireless communication \cite{ref9}, their direct application in wireless link quality time series prediction has not been explored yet. 

Furthermore, multiple parameters across different protocol layers influence link quality in communication systems. Although LLMs perform well in time series forecasting tasks, they face challenges when handling multivariate time series data. This is because LLMs, based on the transformer architecture, typically process one-dimensional time series data \cite{addref18}, whereas multivariate predicting involves multi-dimensional time series data. \cite{addref18} employed dimensional multiplexing to capture the correlations between dimensions by merging multiple time series dimensions into a single sequence before inputting into LLM, thereby improving the multivariate prediction accuracy. However, as the number of time series dimensions increases, the multiplexing and demultiplexing processes become increasingly complex, leading to a degradation in model performance. \cite{addref19} used a multivariate patching strategy to extract correlations among multiple variables, which were then input into the LLM through linear transformations to enhance the performance of multivariate prediction. However, this approach primarily relies on the linear integration of features, limiting its ability to capture complex nonlinear or higher-order relationships. \cite{addref20} proposed a new framework for graph network optimization based on LLM, focusing on optimizing unmanned aerial vehicle (UAV) trajectories and communication resource allocation. Inspired by this, we apply Graph Attention Networks (GAT) to capture the correlations among multiple variables. Unlike \cite{addref20}, we treat the variables as nodes in the graph.

In summary, we propose an innovative model that combines GAT with LLM, termed GAT-LLM. The model enables multivariate prediction of wireless link quality, offering a pathway to intelligent cross-layer optimization in communication systems. The main contributions of this paper are summarized as follows.
\begin{itemize}
\item This study applies LLMs to wireless link quality prediction. By framing link quality prediction as a time series task, we fully exploit the pattern recognition and reasoning capabilities of the LLMs to enhance the accuracy and robustness of wireless link predictions.
\item To address the limitations of LLMs in multivariate prediction, we develop a GAT-LLM model. By constructing multivariate into a graph and using GAT to extract multivariate correlations, this model enhances its capability to handle complex dependencies, thereby improving link quality prediction accuracy and facilitating cross-layer intelligent optimization in wireless networks.
\end{itemize}


\addtolength{\topmargin}{0.04in}
\begin{table}[!t]
\caption{Network Parameters of Dataset}
\begin{center}
\renewcommand{\arraystretch}{1.0}
\resizebox{0.45\textwidth}{!}{
\begin{tabular}{|c|c|c|}
\hline
\textbf{Layer} & \textbf{Parameter Name} &  \textbf{Notes} \\
\hline
\multirow{3}{*}{\raisebox{-1\height}{PHY}} & DLBw &  bps \\
\cline{2-3}
& ULSINR &  db \\
\cline{2-3}
& DLOccupyPRBNum &  \makecell{number, value range 0-100, \\ cell network parameters} \\
\hline
\multirow{3}{*}{MAC} & CellDLMACRate &  bit, cell network parameters \\
\cline{2-3}
& DLMACRate &  bit \\
\cline{2-3}
& MCS &  value range 0-28 \\
\hline
\multirow{3}{*}{PDCP} & PDCPOccupyBuffer &  bit \\
\cline{2-3}
& PDCPUnusedBuffer &  bit \\
\cline{2-3}
& DLPDCPSDUNum &  number \\
\hline
\end{tabular}
}
\label{tab:parameter}
\end{center}
\end{table}

\begin{figure}[t]
\centering
\includegraphics[width=0.35\textwidth]{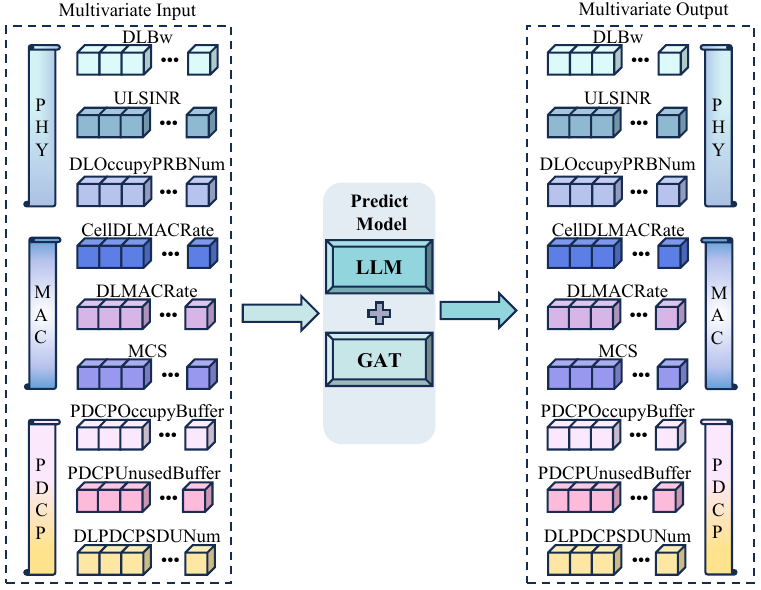}
\caption{Input and output demonstrations of prediction model.}
\label{fig:input-output}
\end{figure}

\begin{figure*}[t]
\centering
\includegraphics[width=16cm]{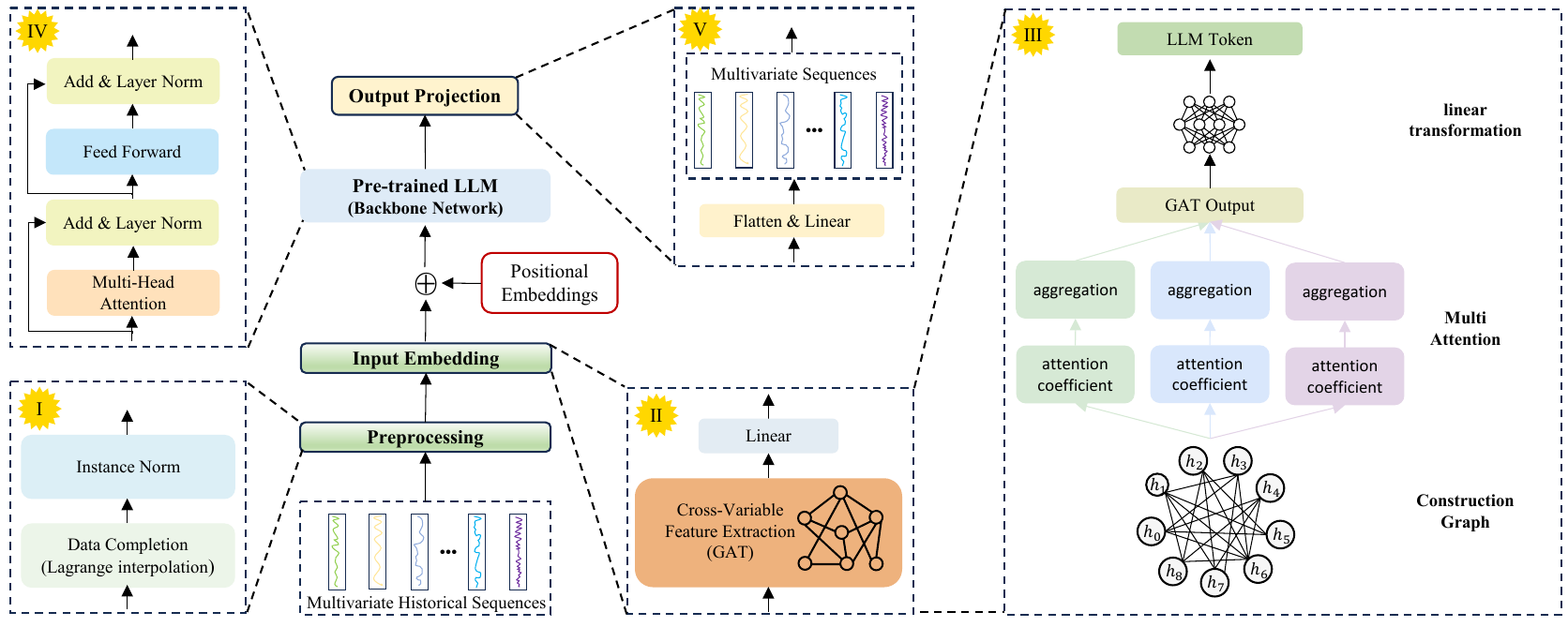}
\caption{The multivariate prediction model framework of GAT-LLM.}
\label{fig:GAT-LLM}
\end{figure*}

\section{Dataset Analysis and Formulation of Multivariate Link Quality Prediction}
\subsection{Dataset Analysis}
The dataset used in this study is the Wireless Link Quality Prediction dataset provided by China Mobile \footnote{https://doi.org/10.12448/3l3e-w818}. With a sampling frequency of 1 ms, the dataset consists of 22,661 data items. Each data item includes nine parameters extracted from the Physical (PHY) layer, Medium Access Control (MAC) layer, and Packet Data Convergence Protocol (PDCP) layer. Specific details regarding these parameters and their associated layers are presented in \Cref{tab:parameter}.

In wireless systems, the quality of wireless links can be assessed through various parameters, each of which will be described in detail below.
\begin{itemize}
\item \textbf{Downlink Bandwith (DLBw)} represents the maximum available data transmission rate from a base station to a user terminal in a wireless communication system. It directly influences link quality by affecting channel utilization and overall network capacity. 
\item \textbf{Uplink Signal to Interference plus Noise Ratio (ULSINR)} refers to the ratio of signal power to noise power when a user device sends a signal to a base station in a wireless communication system. It directly affects signal strength and data transmission rates.
\item \textbf{DLOccupyPRBNum} refers to the number of physical resource blocks (PRBs) occupied in the downlink of a wireless communication system. This metric reflects information, such as resource allocation, making it a critical factor in determining the quality of the downlink.

\item \textbf{CellDLMACRate} denotes the average downlink data transmission rate across the entire cell at the MAC layer, providing a crucial indicator of overall network performance and cell load. This metric is closely associated with link quality.

\item \textbf{DLMACRate} specifically refers to the actual data transmission rate of a user on the downlink, directly reflecting link quality.

\item \textbf{Modulation and Coding Scheme (MCS)} combines modulation techniques with coding strategies in wireless communication systems. It directly impacts the data transmission rate and anti-interference capability. 

\item \textbf{PDCPOccupyBuffer} indicates the occupied buffer capacity in the PDCP layer, which stores data packets awaiting transmission, and \textbf{PDCPUnusedBuffer} denotes the available buffer capacity. They are closely linked to link quality and significantly influence the data transmission success rate and retransmission mechanisms.

\item \textbf{DLPDCPSDUnum} denotes the number of PDCP data units successfully transmitted in the downlink of a wireless communication system. This metric reflects the efficiency and reliability of data transmission and is closely associated with link quality.
\end{itemize}

Based on the preceding analysis, these parameters reflect the link quality from various perspectives. We use the historical time series of these parameters as input for the prediction model, which generates multi-step predictions, as illustrated in \Cref{fig:input-output}.

\subsection{Formulation of Multivariate Link Quality Prediction}
We formulate the prediction of link quality as a multivariate time series problem. In this study, we utilize multivariate, represented by the vector $\boldsymbol{X}^t$, to denote the multidimensional parameters at a specific time $t$ as follows,
\begin{equation}
\boldsymbol{X}^t = \{x_1^t, x_2^t, \cdots, x_N^t\} \subset \mathbb{R}^{T \times N}\;,
\label{vector:X}
\end{equation}
where $T$ represents the duration of the sequence, and $N$ denotes the number of parameters, respectively.

Given the $\mathbb{R}^{T \times N}$ and a fixed window size $\tau$, with $\tau \subset T$, this time series is split into a fixed length input as,

\begin{equation}
\begin{aligned}
\boldsymbol{A} =& \{(\boldsymbol{X}^1, \boldsymbol{X}^2, \cdots , \boldsymbol{X}^{\tau}), (\boldsymbol{X}^{1 + s}, \boldsymbol{X}^{2+s}, \cdots , \boldsymbol{X}^{\tau + s}),\\ &  \cdots , (\boldsymbol{X}^{T-\tau + 1}, \boldsymbol{X}^{T-\tau + 2}, \cdots , \boldsymbol{X}^{T})\}\;,
\label{set:A}
\end{aligned}
\end{equation}
where $s$ denotes the horizontal sliding stride.

When considering the task of predicting the one step value $\boldsymbol{Y}^{t+1}$, the formula is as follows,
\begin{equation}
\boldsymbol{Y}^{t+1} = F(\boldsymbol{X}^{t - \tau + 1}, \boldsymbol{X}^{t - \tau + 2}, \cdots , \boldsymbol{X}^{t})\;,
\label{one-out:Y}
\end{equation}
where ($\boldsymbol{X}^{t - \tau + 1}, \boldsymbol{X}^{t - \tau + 2}, \cdots , \boldsymbol{X}^{t}$) represent the input sequence, while $F(\cdot)$ denotes the prediction model. Furthermore, the predicted value is a multivariate output, defined as follows,
\begin{equation}
\boldsymbol{Y}^{t+1} = \{y_1^{t+1}, y_2^{t+1}, \cdots, y_M^{t+1}\} \subset \mathbb{R}^{1 \times M} \;,
\end{equation}
where $M$ denotes the number of output parameters.

When making multi-step predictions, the autoregressive method is employed to apply formula \eqref{one-out:Y}.

\section{GAT-LLM for Multivariate Wireless Link Quality Prediction}
GAT excels in capturing complex relationships in graph-structured data, while LLMs are proficient in processing complex sequential data. By integrating these approaches, our proposed GAT-LLM model combines the flexibility of graph structures with the sequential modeling capabilities of LLMs, enhancing the accuracy and robustness of multivariate link quality prediction. This model, illustrated in \Cref{fig:GAT-LLM}, comprises four primary modules: preprocessing, input embedding, pre-trained LLM, and output projection.

Initially, multivariate historical data undergoes preprocessing, including interpolation and normalization. The data is then embedded to extract cross-variable features and is linearly transformed. Combined with positional encoding, these embedded sequences are passed through a pre-trained LLM to capture temporal correlations. The output projection module then applies a linear transformation to generate the final multivariate link quality predictions. The following sections detail each module.

\subsection{Input Embedding of the GAT-LLM}
LLMs excel at processing complex sequential data, making them well-suited for time series prediction. However, LLMs face challenges when processing multivariate time series data, as they typically handle one-dimensional data. To address the challenges, we integrate a GAT to capture the correlations among variables, thereby enhancing prediction accuracy. The detailed process is illustrated in \Cref{fig:GAT-LLM} II.

Firstly, a graph $G$ is constructed, comprising nine variables captured at the same moment, where each variable is represented as a node. The learnable weight matrix $\boldsymbol{W}$ is utilized to linearly transform the feature vectors $\boldsymbol{F}$ of all nodes, resulting in a new feature vectors $\boldsymbol{H}$. 


We then calculate the attention coefficient $e_{jk}$ between node $j$ and its neighbor $k$,
\begin{equation}
e_{jk} = LeakyReLU(a^T[\boldsymbol{h_j}][\boldsymbol{h_k}])\;,
\end{equation}
where $a$ is a learnable attention weight vector, $LeakyReLU$ is an activation function, and $[\boldsymbol{h_j}][\boldsymbol{h_k}]$ represents the concatenation of two vectors.

In order to enhance the comparability and stability of the attention coefficient, the softmax function is employed to normalize $e_{jk}$, resulting in the final attention coefficient $\alpha_{jk}$. Then, node $j$ aggregates its feature vector with those of its adjacent nodes through formula \eqref{aggregation}, resulting in a new feature vector $\boldsymbol{h_j^{'}}$,
\begin{equation}
\boldsymbol{h_j^{'}} = \sum_{k \in \mathcal{N}(j)}{\alpha_{jk}\boldsymbol{h_j}}\; ,
\label{aggregation}
\end{equation}
where $\mathcal{N}(j)$ represents the set of neighbors of node $j$, including node $j$ itself.

To enhance the expressive capacity of the model, a multi-head attention mechanism is employed. Each attention head performs the above process independently. Then the output features of each head are connected to get the final feature output,
\begin{equation}
\boldsymbol{h_j^{'}} = ||_{k=1}^K \sum_{k \in \mathcal{N}(j)}{\alpha_{jk}^k\boldsymbol{h_j}^k},
\end{equation}
where $K$ is the number of attention heads, $||$ represents the concatenation operation.

After the multivariate vector $\boldsymbol{X_{i}}$ passes through the GAT layer, the interdependencies between neighboring nodes are captured through linear transformations, attention mechanisms, and feature integration. This process produces a new vector representation, $\boldsymbol{\widetilde{X}_{i}}$, which not only retains the intrinsic features of the node but also incorporates information from other nodes within the graph structure. Consequently, $\boldsymbol{\widetilde{X}_{i}}$ exhibits enhanced semantic richness, making it highly suitable for subsequent predictive tasks, as shown in \Cref{fig:GAT-LLM} III.

Then, we apply a linear transformation on $\boldsymbol{\widetilde{X}_{i}}$ at each moment, converting it into a token that can be recognized by the LLM. Subsequently, we process each input token by positional encoding to enhance the capture of time series information within the sequence.

\subsection{Fine-tuning of the GAT-LLM}
The pre-trained LLMs are primarily designed for text-based tasks, making their direct application to link quality prediction suboptimal. Retraining the entire LLM could address this issue, but the computational costs are prohibitively high. Rather than retraining the entire LLM, we fine-tune all parameters to tailor the model for wireless link quality prediction. In this study, we leverage the general modeling capabilities of LLMs for link quality prediction, using GPT-2 as the backbone. With millions of parameters, GPT-2 provides a robust foundation for the early exploration of LLMs in wireless communication contexts. 

The GPT-2 comprises learnable position embedding layers and stacked transformer decoders, with the number of stacks and feature sizes adjustable as needed. Each layer includes a multi-head self-attention layer, a feed-forward layer, and two layers of normalization layer, as shown in \Cref{fig:GAT-LLM} IV. During the training process, we do not freeze any parameters but rather fine-tuned all parameters. This approach enhances the large language model's suitability for wireless link quality prediction.

\addtolength{\topmargin}{0.04in}
\begin{table}[!t]
\caption{Simulation Parameters}
\begin{center}
\resizebox{0.45\textwidth}{!}{
\begin{tabular}{|c|c|}
\hline
\makebox[0.3\textwidth][c]{\textbf{Parameters}} & \makebox[0.1\textwidth][c]{\textbf{Value}} \\
\hline
Layers & 12 \\
\hline
$n_{embd}$ & 768 \\
\hline
Batch size & 512 \\
\hline
Epochs & 500 \\
\hline
Optimizer & Adam \\
\hline
Learning rate & 0.001 \\
\hline
Hidden dimensions of GAT & 1024 \\
\hline
Number of attention heads of GAT & 16 \\
\hline
Duration of the sequence ($T$) & 22661(ms) \\
\hline
Training step length & 19 \\
\hline
Number of output parameters ($N$) & 9 \\
\hline
Number of output parameters ($M$) & 4 \\
\hline
Fixed window size ($\tau$) & 20 \\
\hline
Horizontal sliding stride ($s$) & 1 \\
\hline
Step size for predicting ($l$) & 10 \\
\hline
\end{tabular}
}
\label{tab:simulation parameter}
\end{center}
\end{table}

\addtolength{\topmargin}{0.04in}
\begin{table*}[tp]
\centering
\fontsize{5.5}{4}\selectfont
\caption{One-step prediction results of GAT-LLM and other benchmark schemes. Black: best.}
\label{one step}
\resizebox{0.75\textwidth}{!}{
\begin{tabular}{c|cc|cc|cc|cc|cc}
	\toprule
	\multirow{3}{*}{\shortstack{\textbf{Multivariate}\\\textbf{Predicting}}}&\multicolumn{2}{c|}{\textbf{GAT-LLM}}&\multicolumn{2}{c|}{\textbf{GPT-2}}&\multicolumn{2}{c|}{\textbf{GAT-Transformer}}&\multicolumn{2}{c|}{\textbf{Conv-LSTM}}&\multicolumn{2}{c}{\textbf{VARIMA}}\cr
	\cmidrule(l){2-3} \cmidrule(l){4-5}\cmidrule(l){6-7}\cmidrule(l){8-9}\cmidrule(l){10-11}
	&MAE&RMSE&MAE&RMSE&MAE&RMSE&MAE&RMSE&MAE&RMSE\cr
	\midrule
    {DLBw}&\textbf{0.0052}&\textbf{0.0073}&0.0098&0.0133&0.0534&0.0774&0.0870&0.1246&0.3163&0.5624\cr
	\midrule
    {ULSINR}&\textbf{0.0029}&\textbf{0.0046}&0.0060&0.0078&0.0232&0.0334&0.0308&0.0419&0.0879&0.2965\cr
        \midrule
    {DLOccupyPRBNum}&\textbf{0.0053}&0.0154&0.0105&\textbf{0.0130}&0.1728&0.2752&0.2346&0.2950&0.2708&0.5204\cr
        \midrule

    {CellDLMACRate}&\textbf{0.0062}&\textbf{0.0081}&0.0095&0.0122&0.0581&0.0775&0.0988&0.1331&0.0994&0.3153\cr
        \midrule
    {DLMACRate}&\textbf{0.0045}&\textbf{0.0071}&0.0072&0.0095&0.0602&0.0870&0.1399&0.1895&0.6398&0.7999\cr
	\midrule
    {MCS}&\textbf{0.0050}&\textbf{0.0070}&0.0081&0.0103&0.0462&0.0802&0.0749&0.1320&0.1135&0.3369\cr
	\midrule    
    {PDCPOccupyBuffer}&\textbf{0.0028}&\textbf{0.0043}&0.0061&0.0084&0.0213&0.0293&0.0331&0.0445&0.0844&0.2905\cr
        \midrule	
    {PDCPUnusedBuffer}&\textbf{0.0026}&\textbf{0.0053}&0.0042&0.0064&0.0183&0.0275&0.0229&0.0315&0.2349&0.4847\cr
        \midrule
    {DLPDCPSDUNum}&\textbf{0.0045}&\textbf{0.0073}&0.0066&0.0087&0.0570&0.0816&0.1288&0.1752&0.1779&0.4218\cr
	
	\bottomrule
\end{tabular}
}

\end{table*}

\begin{figure*}[t]
    \centering
    \begin{minipage}[b]{0.4\textwidth} 
        \centering
        \includegraphics[width=\textwidth]{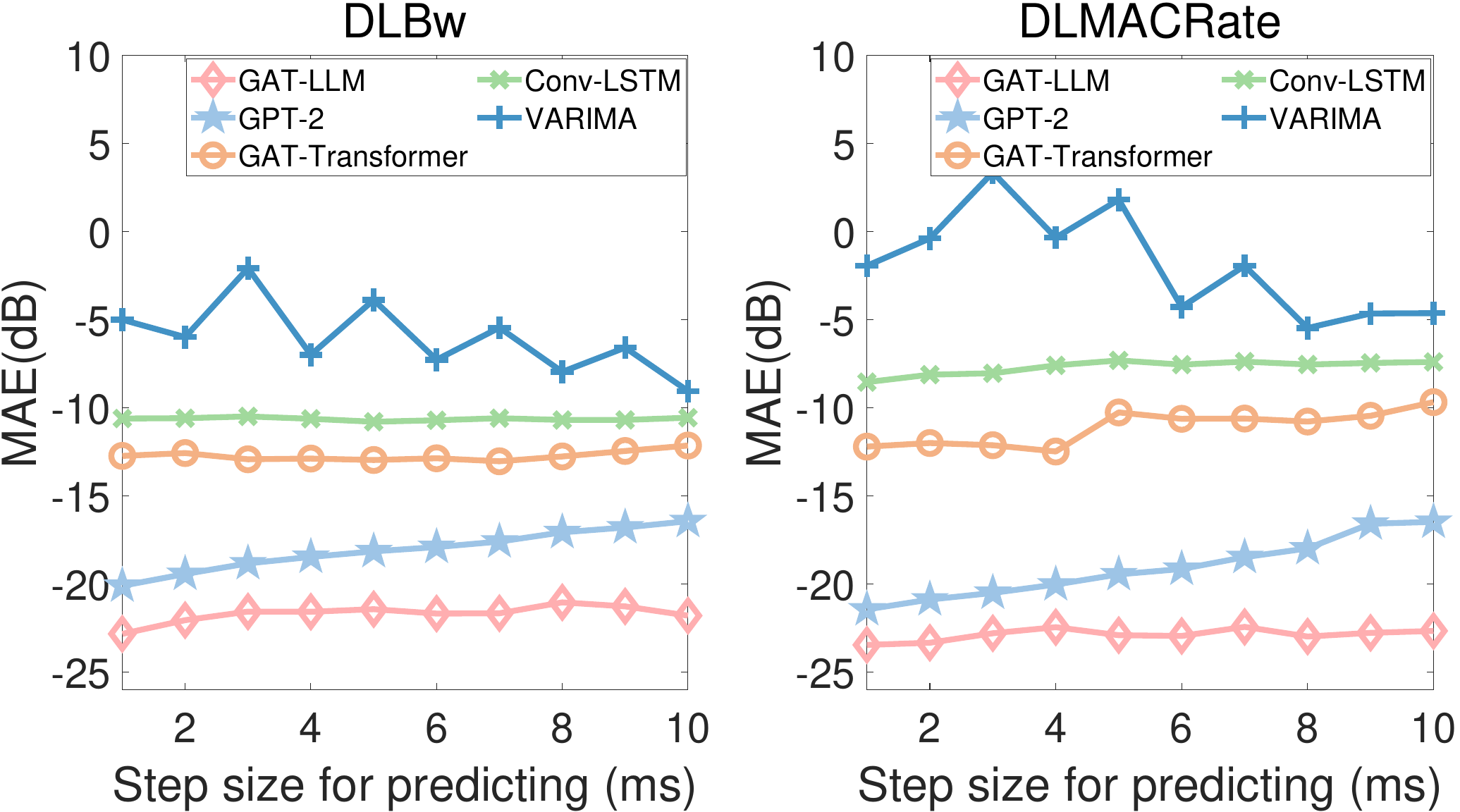} 
    \end{minipage}
    \hspace{0.01\textwidth} 
    \begin{minipage}[b]{0.4\textwidth} 
        \centering
        \includegraphics[width=\textwidth]{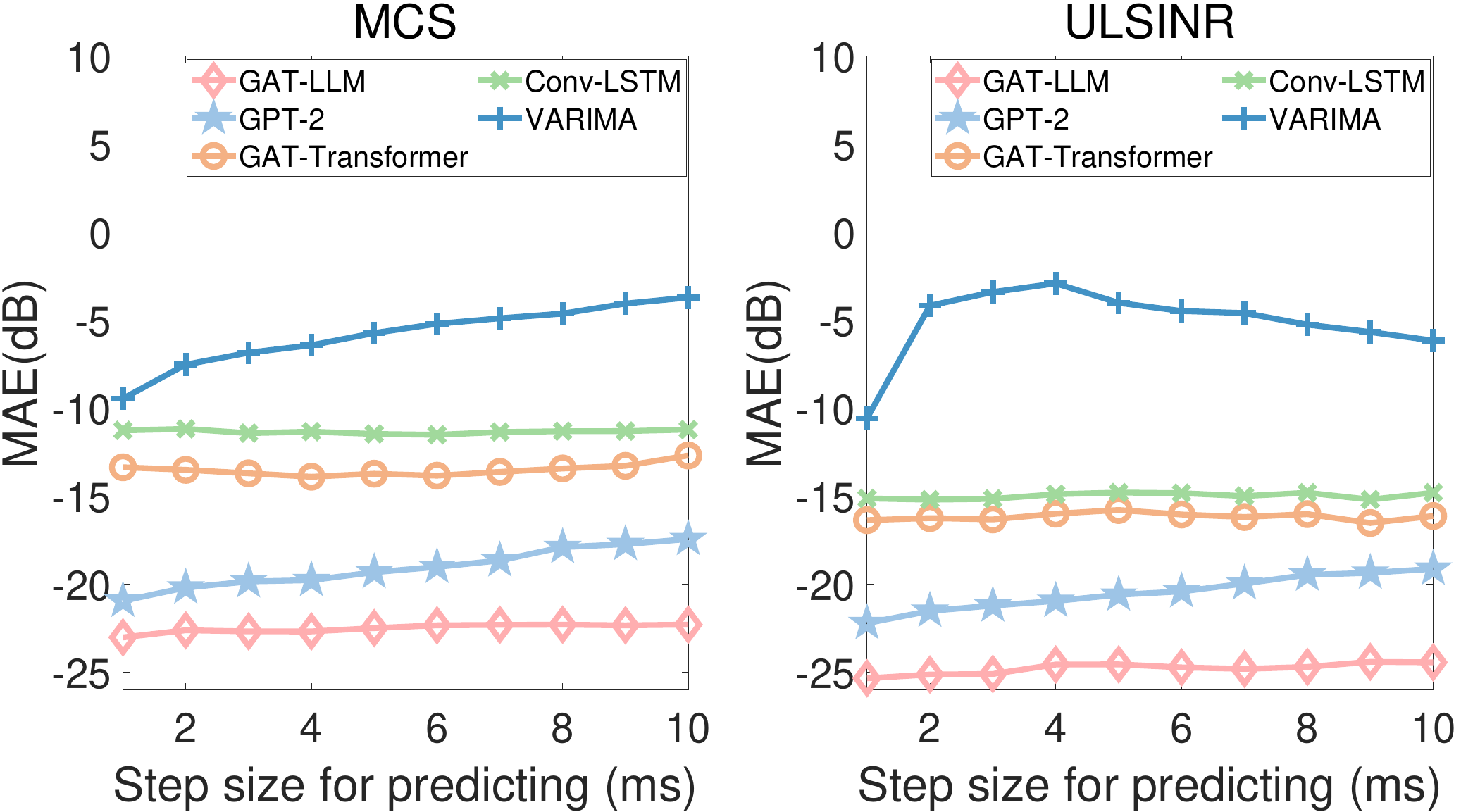} 
    \end{minipage}
    \caption{Comparison of multivariate link quality prediction performance among GAT-LLM, GPT2, GAT-transformer, Conv-LSTM and VARIMA.}
    \label{fig:new_vs_benchmark}
\end{figure*}

\subsection{Input and Output of the GAT-LLM}
The part includes two parts: preprocessing input and output projection.
\begin{itemize}
\item \textbf{Preprocessing:} 
In wireless systems, the instability of wireless channels frequently leads to incomplete data collection \cite{ref2}, as is the case with the dataset used in this study. To address the issue of missing data, we employ Lagrange interpolation polynomials. Additionally, we apply min-max normalization to scale all parameters linearly to [0, 1], reducing discrepancies between different features in the dataset, as shown in \Cref{fig:GAT-LLM} I.


\item \textbf{Output Projection:}
The output representations generated by the pre-trained LLM are flattened and subjected to a linear projection to produce the final prediction, denoted as $\boldsymbol{Y_{i}}$. In the output projection module, GAT-LLM employs a non-linear fully connected layer to map the tokens to wireless link quality data, as shown in \Cref{fig:GAT-LLM} V.
\end{itemize}

\section{Experiments}
\subsection{Experimental Settings}

\begin{enumerate}
\item Evaluation Metrics

We evaluate the performance of the proposed framework using three common metrics: Mean Absolute Error (MAE) and Root Mean Square Error (RMSE). The formula is as follows,
\begin{equation}
MAE = \frac{1}{n} \sum_{i=1}^{n}{|y_i - \hat{y}_i|}\; ,
\end{equation}
\begin{equation}
RMSE = \sqrt{\frac{1}{n} \sum_{i=1}^{n}{(y_i - \hat{y}_i)^2}}\; ,
\end{equation}

where $y_i$ represents the true value, $\hat{y}_i$ denotes the predicted value,  $\bar{y}$ represents the mean of the actual values and $n$ indicates the number of samples.

Smaller MAE and RMSE values indicate better model performance.


\item Simulation Setup

We choose GPT-2 as the backbone model due to its effective trade-off between inference speed and prediction accuracy. Notably, our method is theoretically applicable to other LLMs. Additionally, we employ GAT for cross-variable feature extraction, with specific parameter details provided in \Cref{tab:simulation parameter}. The GAT-LLM model was implemented using PyTorch 2.0.1, and all experiments were conducted on a server equipped with 2 5318Y CPUs and 2 NVIDIA RTX 4090 GPUs. This model is intended for future deployment on the base station side, which is feasible given the base station's computational capabilities.
\end{enumerate}

\subsection{Prediction Performance Analysis}
To evaluate the effectiveness of the GAT-LLM model for multivariate wireless link quality prediction, we compare its performance with four benchmark schemes: GPT-2, GAT-Transformer, Conv-LSTM, and VARIMA. The specific implementation details of the GAT-LLM model are provided in Section III.

\begin{itemize}
\item \textbf{GPT-2 \cite{ref6}:} The GPT-2 is an autoregressive language model built on the Transformer architecture, known for its proficiency in processing time series data and capturing long-range dependencies. 

\item \textbf{GAT-Transformer \cite{ref13}:} The GAT-Transformer model is a relatively recent scheme that combines GAT with the Transformer architecture. The scheme excels in capturing multivariate relationships and long-range dependencies in time series data. 

\item \textbf{Conv-LSTM \cite{ref3-1}:} Conv-LSTM integrates the strengths of Convolutional Neural Networks (CNNs) and LSTM networks. It is particularly well-suited for handling spatiotemporal sequence data, as it effectively captures both spatial and temporal dependencies in the data.

\item \textbf{VARIMA \cite{ref-arima}:} The Autoregressive Integrated Moving Average (ARIMA) model is a statistical method used for time series analysis and prediction. To support multivariate predicting, we employ its multivariate extension, Vector ARIMA (VARIMA), designed to handle multiple interrelated time series.

\end{itemize}

\begin{figure*}[t]
    \centering
    \begin{minipage}[b]{0.2\textwidth} 
        \centering
        \includegraphics[width=\textwidth]{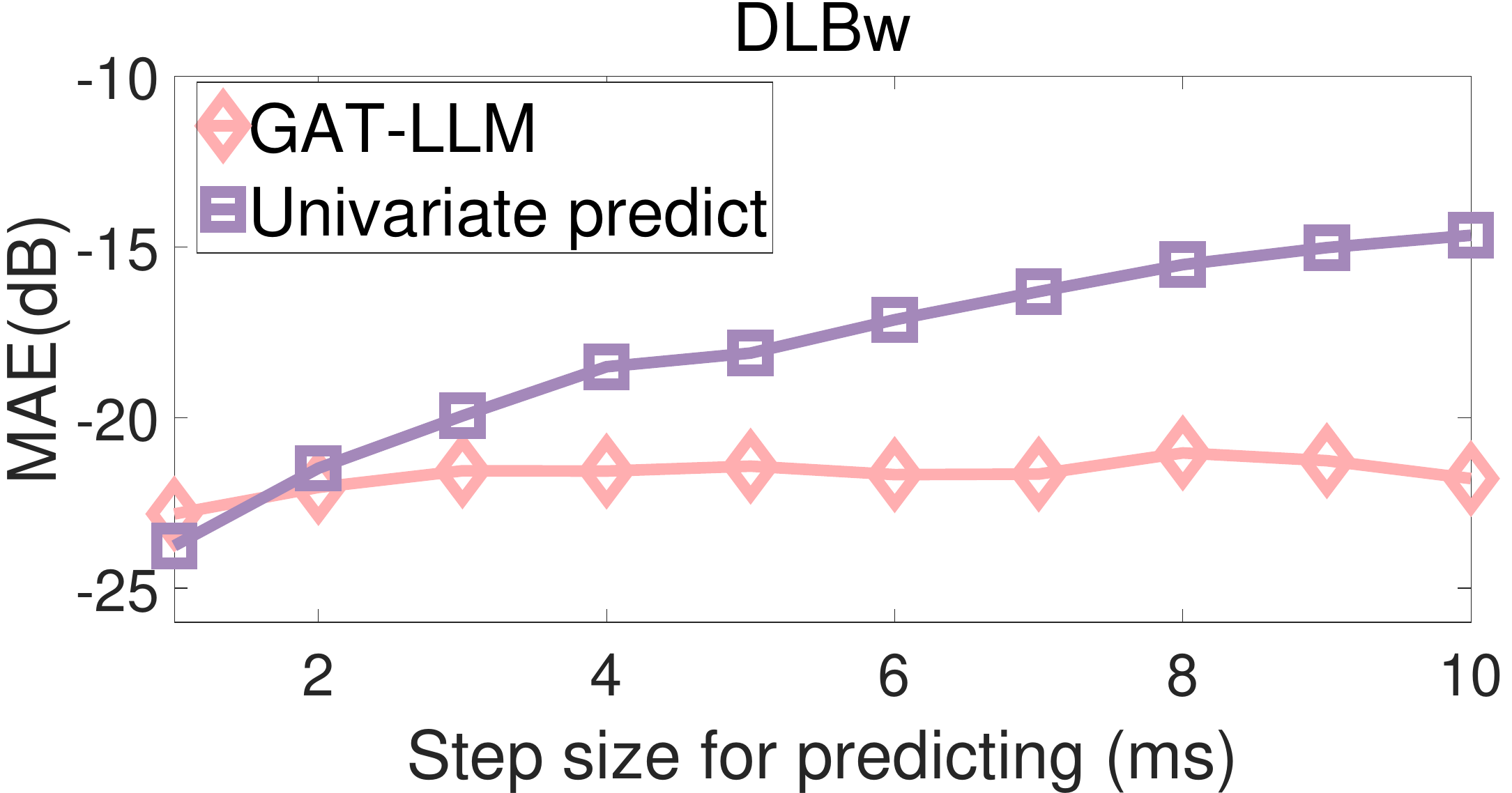} 
    \end{minipage}
    \hspace{0.01\textwidth} 
    \begin{minipage}[b]{0.2\textwidth} 
        \centering
        \includegraphics[width=\textwidth]{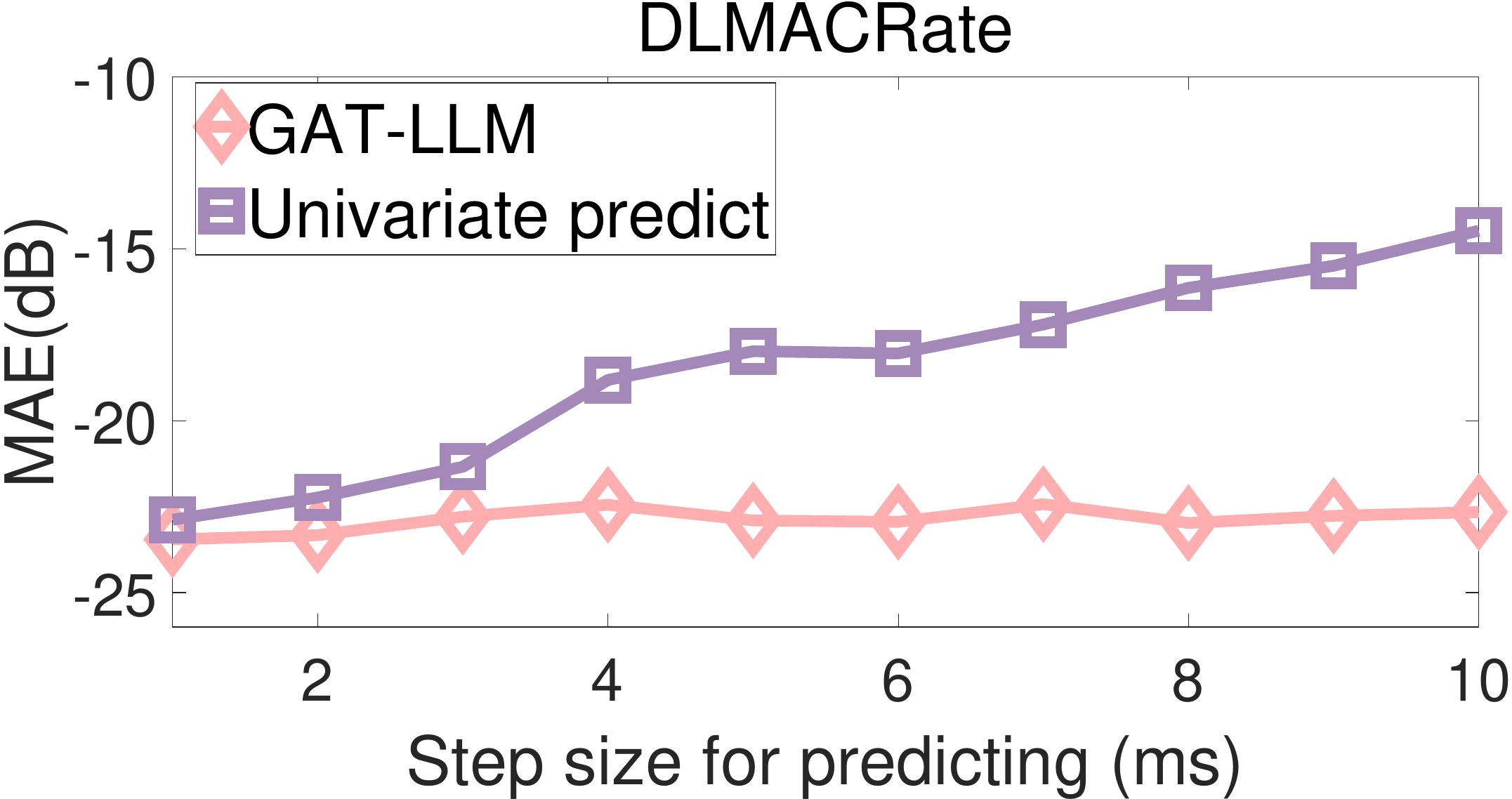} 
    \end{minipage}
    \hspace{0.01\textwidth} 
    \begin{minipage}[b]{0.2\textwidth} 
        \centering
        \includegraphics[width=\textwidth]{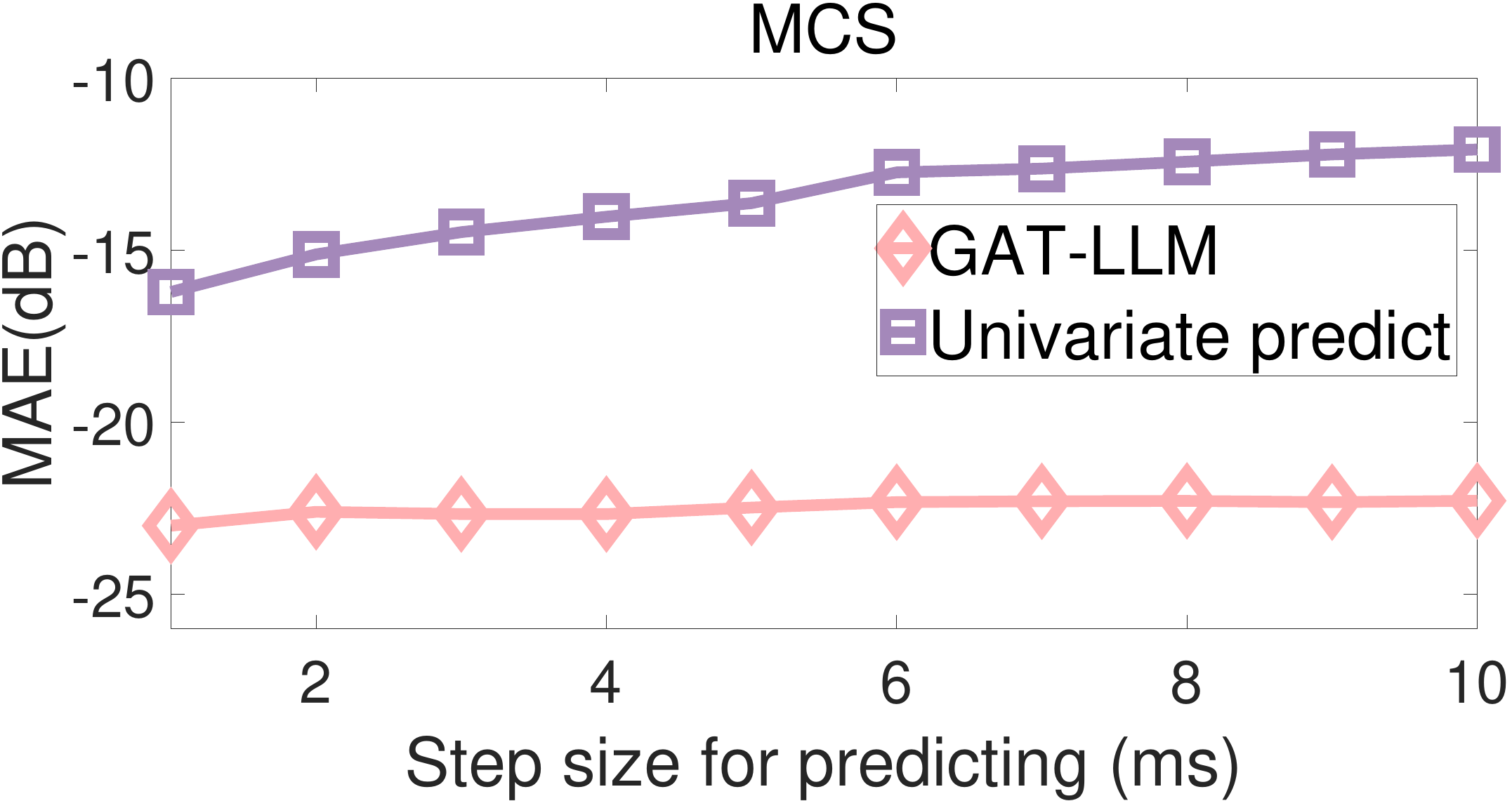} 
    \end{minipage}
    \hspace{0.01\textwidth} 
    \begin{minipage}[b]{0.2\textwidth} 
        \centering
        \includegraphics[width=\textwidth]{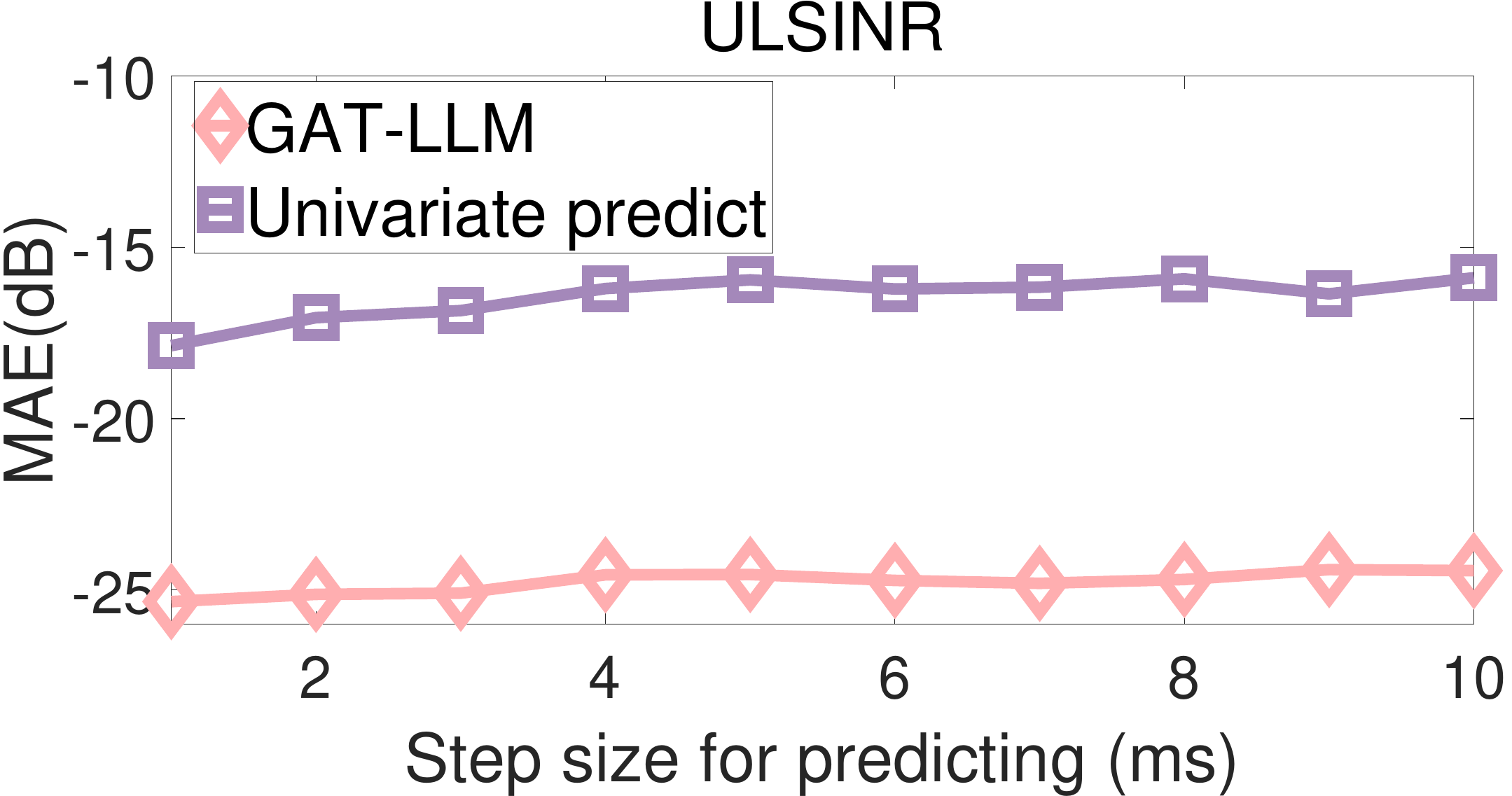} 
    \end{minipage}
    \caption{Comparison of link quality prediction performance between Univariate
predict and GAT-LLM.}
    \label{fig:new_vs_univariate}
\end{figure*}


To ensure the fairness of the experiments, the models of all benchmark schemes, except VARIMA, are configured with consistent parameters such as hidden layer dimension and number of layers. \Cref{one step} presents the MAE and RMSE results for the one-step prediction across the five schemes. As shown, GAT-LLM outperforms all other models in predicting link quality across most variables, with two exceptions: DLOccupyPRBNum, which performs worse than the GPT-2 scheme in terms of RMSE. These results demonstrate the unique advantages of the proposed GAT-LLM scheme in handling multivariate link quality prediction.

Further, we perform 10-step prediction simulations, with each step corresponding to 1 ms. Due to the large number of prediction parameters, we select four key parameters, DLBW, DLMACRate, MCS, and ULSINR, for performance demonstration. Specifically, DLBW and DLMACRate capture the downlink bandwidth and transmission rate, respectively; MCS reflects the modulation and coding characteristics of the physical layer, and ULSINR measures the uplink signal-to-noise ratio. These parameters provide a comprehensive link quality assessment from the physical to the transport layer, establishing a solid foundation for accurate prediction and analysis. \Cref{fig:new_vs_benchmark} compares the performance of the proposed GAT-LLM scheme against other benchmark schemes in multi-step link quality prediction. Firstly, the results demonstrate that the GAT-LLM scheme outperforms the other schemes, which can be attributed to its integration of GAT and LLM, effectively addressing the complexities of multivariate link quality prediction. Additionally, we observe that GPT-2 is the suboptimal scheme. This success can be explained by its Transformer-based architecture, which excels at capturing long-range dependencies. After fine-tuning, GPT-2 proves to be well-suited for temporal link quality prediction tasks. Furthermore, all deep learning-based schemes outperform the VARIMA scheme. This is because deep learning schemes excel at capturing intricate nonlinear relationships and automatically extracting relevant features, thereby improving prediction accuracy. In contrast, the VARIMA scheme relies heavily on linear assumptions, which limits its ability to handle complex, multivariate data. 

\subsection{Comparison Between Univariate and Multivariate}
The primary advantage of the GAT-LLM model lies in its capacity to predict multivariate time series data. To assess the impact of multivariate data on prediction accuracy, we compared the performance of GAT-LLM with a univariate predicting scheme. In the univariate prediction scheme, the predicting model is the GAT-LLM, but its input is only the historical data of the target variable.

\Cref{fig:new_vs_univariate} shows a performance comparison between the GAT-LLM scheme and a univariate prediction scheme. While the univariate scheme performs marginally better in the first step of DLBW prediction, GAT-LLM demonstrates a significant advantage as the prediction horizon increases. This superiority is due to GAT-LLM’s ability to capture interdependencies across multivariate, providing a more comprehensive information base for future predictions, which is an advantage that the univariate prediction scheme lacks. These results highlight the critical role of multivariate prediction in improving long-term prediction accuracy.



\section{Conclusion}
In this paper, recognizing the critical role and challenges of wireless link quality prediction, we propose an innovative link quality prediction model that integrates LLMs and the GAT, referred to as GAT-LLM. We reformulate the link quality prediction task as a time series problem to leverage LLMs' capabilities in handling sequential data. Recognizing that link quality is influenced by multiple interdependent parameters, we employ GAT to capture correlations among these variables. The GAT-LLM model thus combines GAT's strength in modeling multivariate relationships with LLMs' ability to process rich contextual and historical information, effectively addressing the complexities inherent in multivariate link quality prediction. Experimental results demonstrate that GAT-LLM enhances the accuracy and robustness of link quality predictions, particularly in multi-step predicting scenarios. It indicates that GAT-LLM holds significant potential as a valuable tool for tackling the complexity of wireless link prediction in real-world communication systems.



\bibliographystyle{IEEEtran}
\renewcommand{\bibfont}{\footnotesize}
\bibliography{ref.bib} 

\end{document}